%% file: main.tex
\documentclass[10pt]{article} 
\usepackage[preprint]{tmlr}

\input{math_commands.tex}

\usepackage{hyperref}
\usepackage{url}
\usepackage{booktabs}
\usepackage{graphicx}
\usepackage{capt-of}

\title{What It Costs to Compose, Rebuild, and Correct Precomputed Memory}

\author{\name Asa Shepard \email as66@williams.edu \\
      \addr Department of Computer Science,
      Williams College
      }

\def\month{MM}
\def\year{YYYY}
\def\openreview{\url{https://openreview.net/forum?id=XXXX}}

\begin{document}
\maketitle

\begin{abstract}
Language models can answer from precomputed memory, a model's saved reading of a body of material, reused across requests instead of read again at each. This paper maps where that practice preserves correctness and the conditions under which it fails. Across experiments on Llama-3.1-8B-Instruct using both saved key-value caches and trained compressions of them, precomputed memory degrades when assembled from separately prepared parts, stays current only through rebuilds costing a large fraction of full preparation in our measurements, and ignores corrections served beside it conditional on phrasing. If precomputed memories can be served alongside one another, be cost-efficiently rebuilt, and be superseded by new information arriving in real-time, they can serve as a way to avoid re-feeding context to a model over repeated queries. The implication of our results for a deployed system that deals with a variety of queries is that precomputed memories are best rebuilt on the cadence at which new information changes what the memory was originally computed from. Both warm-rebuilding trained compressions of key-value caches and serving specifically-phrased updates beside a memory, as pasted text or injected cache state, show particular promise for keeping precomputed memories current, the latter as an interim measure between rebuilds, and we measure the cost and name the remaining questions associated with each.
\end{abstract}

\vspace{-8pt}
\noindent\begin{minipage}{\linewidth}
\setlength{\abovecaptionskip}{4pt}
\centering
\includegraphics[width=\linewidth]{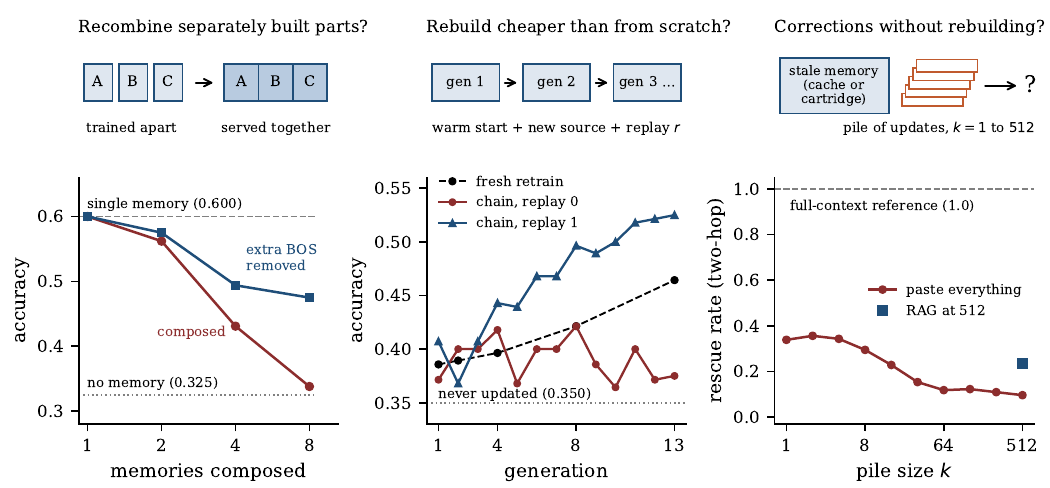}
\captionof{figure}{The three questions. Left, build time: separately trained memories are concatenated at serving time, and accuracy falls with the number of memories composed. Removing the duplicated start-of-text columns recovers about half the fall (Section~\ref{sec:compose}). Middle, build time: one evolving memory is rebuilt warm as sources accumulate, and replay of the earlier sources' training data is what separates chains that track a fresh retrain from chains that decay to the never-updated floor (Section~\ref{sec:warmstart}). Right, query time: an already-built memory is served together with a pile of updates. On two-hop items, use of the correction falls as revisions accumulate, and no delivery policy comes within 64 points of the full-context reference (Section~\ref{sec:corrections}).}
\label{fig:overview}
\end{minipage}
\vspace{1em}

\section{Introduction}

We test two different forms that a model's saved state can take, in the setting where a stream of diverse queries arrives against a common corpus, such as a legal case, a medical profile, or company records. The default way to serve such queries is in-context learning \citep[ICL;][]{brown2020gpt3}. The material is placed in the model's context window and the model conditions on it, paying a \emph{prefill}, the model's single forward pass over the input, which computes and stores each token's key and value vectors before any output is generated. A \emph{cache} is that key-value state exactly as one prefill pass produces it, ICL's internal representation of the material saved for reuse. It's cheap to make (one forward pass), contains one stored entry per token, and thus it's as large in storage as the text it came from. A \emph{cartridge} is a smaller cache trained to imitate the full one \citep{eyuboglu2025cartridges}. It's expensive to make, but it's compressed, with the training cost meant to be amortized over every future query against the same material. We call either a precomputed memory, and we test both to help us determine whether a limit belongs to one substrate or to precomputed memory as a whole, as well as to test the more versatile form of cartridges, which are reported to deploy with a smaller footprint over larger amounts of context without sacrificing accuracy.

The key and value vectors stored for any given token that has passed through a model are computed with attention over everything else present in the same pass. For example, token 1001 requires information from tokens 1--1000, so if any of those first 1000 tokens are changed, then token 1001 will be stored differently in the model's key-value cache. A cartridge inherits the same property, and uses synthetic conversations about the corpus of data it is ``memorizing'' to train itself, a method called self-study.

New information can enter a memory at two points (Figure~\ref{fig:overview}). It can be present at \emph{build time}, in the prefill or training data while the memory is made, or it can arrive at \emph{query time}, served next to the finished memory as prompt text or as state written into the cache. For example, a memory about Bob's finances is built in January, and in March, Bob buys a new house. The build-time option is to rebuild the precomputed memory from scratch but now including the new fact that Bob bought a house, and the query-time option is to hold the original January memory static and supply the new March information alongside it. With precomputed memories, the model only has to attend over the memory's $X$ static tokens about Bob's financial history at every pass. Training a cartridge is more expensive up front, but it compresses that further, so the model only has to attend over $X/c$ tokens, for a compression ratio $c$, with each new request. However, this is only a clear benefit in a world where the memory's information is never updated or superseded, since a corpus that never changes and is queried enough amortizes the cost of any build, however expensive, without risking a staleness-caused decline in accuracy.

\begin{figure}[t]
\begin{center}
\includegraphics[width=\linewidth]{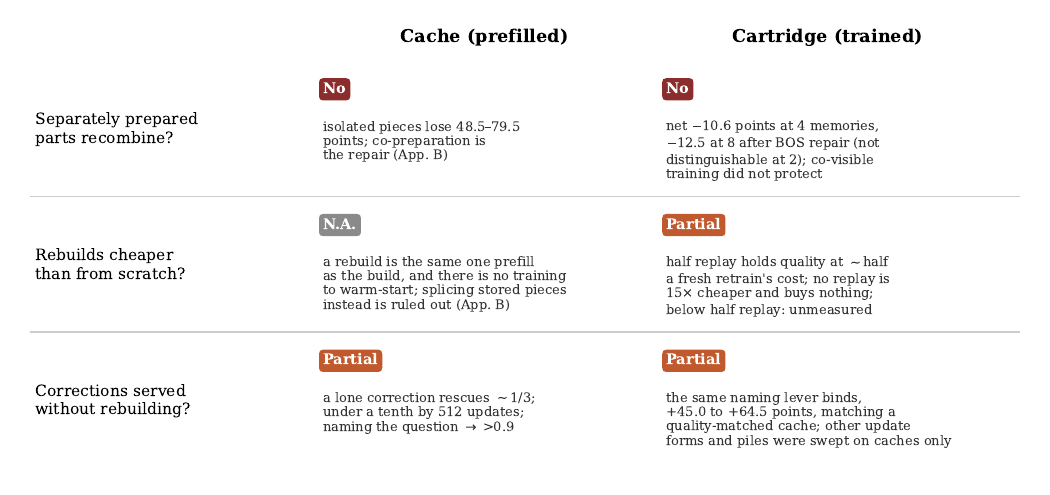}
\end{center}
\caption{The three questions of the introduction, answered per substrate on this model. Sections~\ref{sec:compose} through~\ref{sec:corrections} carry the measurements. The premise that preparation cannot be made partial, which is why full joint preparation is treated as given throughout, is measured in Appendix~\ref{app:threshold}.}
\label{fig:questions}
\end{figure}

Whether the idea scales past that depends on three questions. Can separately prepared parts be recombined, enabling preparation to only be done once per block of information? Can rebuilds cost less than building from scratch, making updates to existing memories more cost-effective? And can corrections be served without rebuilding at all, letting a memory stay static between rebuilds? A prior question, whether preparation can be made partial, has its answer predicted by prior work on modular context \citep{ratner2023pcw, yang2024revisitingpcw, yang2025ape} and confirmed at measured magnitudes in Appendix~\ref{app:threshold}: A cache delivers 95\% of its benefit on questions that connect facts only once 82--92\% of its text was visible during preparation, a constant fraction across a 16$\times$ range of length (fitted slope $-0.045$, 95\% CI $-0.130$ to $+0.046$, replicated on fresh items), so throughout the paper, full joint preparation is treated as necessary per-unit. The three questions each have their own section. The answers are the paper's contributions, and Figure~\ref{fig:questions} previews them per substrate.

\begin{enumerate}
  \item \textbf{A composition penalty for cartridges and a method for partial recovery.} Composing separately-trained memories costs accuracy. We attribute about half of the loss we observe with composing separately-trained cartridges to duplicated start-of-text markers, a mechanical artifact of concatenation, and show that the remainder survives partially co-visible training (Section~\ref{sec:compose}). The result sits between the composability reported by \citet{eyuboglu2025cartridges} and the collapse reported by \citet{hardalov2026cas}.
  \item \textbf{Warm-start retraining, with replay as the controlling factor of both quality and cost.} Rebuilding a cartridge memory warm as sources accumulate stays within a workable distance (within ten points, the smallest gap the corpus we use can detect) of fresh retraining through the deepest chain our corpus supports, but only with replay of the older material, mixing the earlier sources' training data back into each rebuild (Section~\ref{sec:warmstart}). The no-replay chain is 15$\times$ cheaper per rebuild and buys nothing, with thirteen successive updates worth nothing measurable over never updating at all. The full-replay chain preserves quality at slightly more than a fresh retrain's cost, and the half-replay chain matches full replay within the detectable margin at about half the cost of a fresh retrain.
  \item \textbf{The query-time collapse, decomposed.} At query time, when a lone realistic correction to a precomputed memory (a prefilled cache, or the stale text re-read) is served beside the memory, the correction is used (repeated correctly when the model is asked a question about it) on about a third of a set of two-hop questions (questions that require reasoning over two independent facts at once), whether pasted as text or injected into the cache. Use falls as further revisions accumulate to under a tenth at 512 (Section~\ref{sec:corrections}). No policy for delivering updates on top of a static memory that we measured is workable on our configuration, with every cell at least 20 points short of the full-context reference and the two-hop cells at least 64 short. The accumulation cost is the update's content rather than the size of the update in tokens, because token-matched padding that revises nothing costs far less in accuracy than real revisions do, a contrast that replicates on fresh items. And swept over all ten pile sizes, we only observe the cost with two-hop questions, while single-hop questions gain 16.7 points from updates kept in context before plateauing near 32.
  \item \textbf{Naming the question is what makes an update bind.} An update that names its question is used on over nine-tenths of items, a labeled value that names nothing is used on a tenth, and whether the delivery mechanism is pasted text or cache injection changes the outcome by less than a point (Section~\ref{sec:corrections}). The same binding occurs with trained cartridges. The effect replicates on fresh items, and a wrong-naming control locates its mechanism not as recognition, but as compliance, since a false value that names the question is adopted nearly as readily as the true one.
\end{enumerate}

\section{Related work}

\subsection{Precomputed and trained context representations}

Cartridges \citep{eyuboglu2025cartridges} distill a body of text into a small trained key-value cache by self-study. During the process of self-study, the model generates synthetic conversations about the material, and a cache is trained with an objective to reproduce the model's behavior as if the full text were in context. Two of that paper's claims matter here. First, training the small cache by plain next-token prediction on the material performs far worse than simply placing the material in context, so self-study is required. Second, cartridges trained on different texts can be loaded side by side and queried across, without any further training, a claim demonstrated on pairs of cartridges. Cartridges at Scale \citep{hardalov2026cas} contradicts the second claim for the naive case at larger counts, reporting that mixing cartridges trained in isolation collapses accuracy to near chance, 26.0\% on 5-way multiple choice at 20 cartridges, and proposes mixed-visibility training, in which each cartridge trains with material from a sampled number of other sources in view, recovering accuracy to 77.8\%. Our composition experiments sit between these two claims, and Section~\ref{sec:compose} states what we can and cannot adjudicate. We adopt the published self-study recipe unmodified and measure preparation and update limits of the existing techniques on our configuration.

\subsection{Knowledge conflict and updating}
\label{sec:conflict}

The query-time failure we measure is a knowledge-conflict behavior in which the model reads a stale text and often repeats the outdated fact while the correction sits in the same context, connecting it to work on entity-based knowledge conflicts and context faithfulness under contradiction \citep{longpre2021entity}. It also connects to parameter-level knowledge editing \citep{meng2022rome, meng2023memit}, where updates are written into parametric memory, the weights, and where edited facts often fail to propagate through multi-hop questions \citep{zhong2023mquake, cohen2024ripple}. Low-rank adapter patching \citep{hu2021lora} is the corresponding parametric alternative, and an internal precursor measurement of it appears in Appendix~\ref{app:precursor}. The appeal of not using the weight pathway is best put in the framing of \citet{eyuboglu2025cartridges}. A cartridge is prefix tuning \citep{li2021prefix}, the deeper-layer cousin of prompt tuning \citep{lester2021power}, extended to long contexts, so its content enters through attention exactly as text does, which supports chaining in a way weight edits often do not, and the published comparison finds the prefix parameterization beating memory-matched LoRA both in and out of domain while serving without adapter-specific infrastructure. Likewise, our results show that pasted text and injected state behave alike at a single update, and whether the update's own wording names the question it bears on more heavily influences whether new updates are used (Section~\ref{sec:corrections}). The wrong-naming control of Section~\ref{sec:corrections} sharpens the connection: On our configuration, an update that names a question is adopted nearly as readily when its value is false, so an update channel behaves as an injection surface in the sense of the indirect-prompt-injection literature \citep{greshake2023injection} and of concurrent attacks on agent memory, where planted entries steer later answers and content screening fails to catch them \citep{tian2026injecmem, karunanidhi2026poisoning}, and control of update wording is control of the answer. The pull of text that names the question is itself an old observation, since distractor sentences constructed to overlap the question fooled extractive reading-comprehension models in the same way \citep{jia2017adversarial}.

\subsection{Retrieval over updates}
\label{sec:rag}

The failure this paper measures is downstream of retrieval. Retrieval-augmented generation \citep{lewis2020rag} would seem to answer the staleness problem by retrieving the relevant update and placing it in context. Our retrieval arm is lexical BM25 \citep{robertson2009bm25} over the update pile with one round of pseudo-relevance feedback \citep{croft1979probabilistic}, using Rocchio-style query expansion \citep{rocchio1971relevance}. Pseudo-relevance feedback pulls linking passages for two-hop questions in from the base text, and the expanded query ranks the correct update first at every pile size tested. The retrieval arm is therefore effectively gold retrieval, and it is the strongest policy we test at large piles. On the two-hop items, it still ends at least 64 points short of the full-context reference, though that may also be influenced by the baseline reasoning capability of the model.

\section{Reading the results}
\label{sec:setup}

Except where a second model is named, every number comes from Llama-3.1-8B-Instruct \citep{grattafiori2024llama3} in bf16, with fully deterministic decoding, so replication throughout this paper refers to fresh items rather than reruns, and all verdicts are claims about this model. A descriptive check on a second model appears in Section~\ref{sec:corrections}, and precursor measurements, reported for grounding only, appear in Appendix~\ref{app:precursor}. Hardware, decoding, chat-template handling, and measured instrument noise are in Appendix~\ref{app:setup}.

\paragraph{Two substrates.} Throughout, caches are prefilled and cartridges are trained, never the other verb, and Section~\ref{sec:discussion} collects the head-to-head evidence.

\paragraph{Texts and questions.} The synthetic experiments use texts of roughly 1{,}500 to 24{,}000 tokens, sequences of short record-keeping paragraphs about invented people, projects, and places, constructed so that no fact is answerable from pretraining. Appendix~\ref{app:setup} gives the construction. The cartridge experiments train and evaluate on patient records from the LongHealth benchmark of long clinical texts \citep{adams2024longhealth}. Questions come in two tiers. A \emph{single-hop} question asks for a fact stated in one paragraph. A \emph{two-hop} question requires combining two separately stated facts. Every synthetic question has exactly one or two hops; the clinical records' questions are not hop-labeled (Section~\ref{sec:discussion}). Distractors, the wrong options offered beside the right answer, come in two levels of distractor difficulty (how similar they are to the relevant fact(s), explained in Appendix~\ref{app:measure}), and answers are collected in both multiple-choice and free-form formats.

\paragraph{How results are scored.} The upper reference throughout is the \emph{full-context read}, in which the model reads the entire current text in the prompt. We should note that the full-context read is deployable only to the point that the context fits inside the model's context window, and as the amount of context reaches that level, performance degrades \citep{liu2024lost, hong2025context}. Cartridges work toward bypassing that limitation. A \emph{floor} is measured accuracy with no relevant text in view, and it can sit below nominal chance because items answerable without the text are screened out in advance. The correction experiments screen further, keeping only items the model answers correctly from the corrected text and incorrectly from the stale one, so their floor is 0 and their ceiling is 1 by construction. Accuracy on that scale is a \emph{rescue rate}, the fraction of guaranteed failures a policy repairs, and it is not comparable to ordinary accuracy. Appendix~\ref{app:setup} states the rules in full, including the minimum detectable effect (MDE) convention used throughout, and records which were registered in advance and which are conventions adopted during analysis, and Appendix~\ref{app:measure} measures how much evaluator choices move the numbers reported here.

\section{Build time: composing separately built memories}
\label{sec:compose}

\subsection{Motivation}

This section takes the question: Can separately prepared parts be recombined? This section is about cartridges, and the piece experiment of Appendix~\ref{app:threshold} is its cache-side counterpart. The appeal of one memory per source text is composition on demand. \citet{eyuboglu2025cartridges} report that self-study cartridges compose at inference time, while \citet{hardalov2026cas} report that naive composition collapses to near chance at 20 cartridges and that training each cartridge with other sources' material in view repairs it. We investigate how much composition costs once assembly mechanics are controlled for, and whether partially co-visible training protects against the penalty.

\subsection{Setup}

We first validate the substrate. The published self-study recipe is implemented unmodified, including its \emph{seed prompts}, the five generic prompt types from which self-study conversations are generated. The recipe keeps them deliberately corpus-agnostic \citep{eyuboglu2025cartridges}, and tailoring them to a specialized workload is a lever we likewise leave untouched (Section~\ref{sec:discussion}). Cartridges are trained across a sweep of memory budgets on LongHealth patient records, and each is compared against its own initialization at two independent seeds. Self-study duration, the recipe's own term for training length, is set here by the generated data. Self-study yields roughly 6{,}200 to 8{,}100 synthetic conversations per record, training material never includes evaluation items, and metered training costs are in Appendix~\ref{app:setup}.

For composition, 2, 4, or 8 cartridges are placed at consecutive position blocks (each cartridge's columns occupy the next contiguous run of position indices, with no interleaving) and concatenated, and accuracy is compared against a single-cartridge baseline, in which each item is scored against the cartridge trained on that item's own source record.

One mechanical difference between a composed cache and a normal one is that each separately-built cartridge includes its own frozen beginning-of-sequence (BOS) column, the start-of-text token that a Llama-3.1 context contains exactly once at position zero and that serves as an attention sink \citep{xiao2023sinks}, so an 8-cartridge memory contains 8 BOS columns where a normal context has 1. The \emph{start-marker control} isolates this defect by evaluating the same composed memory with the extra BOS columns removed, keeping only the first BOS column. We run two further checks: removing an ordinary trained column instead, which keeps cache length and positions identical and which tests whether any deletion helps, and removing every BOS column including the first, which tests whether the marker matters at all.

For our co-visible training conditions, 30\% of each cartridge's self-study conversations draw their source material from texts other than the cartridge's own. Two arms hold that fraction and the total co-visible token budget identical and vary only how many distinct other sources supply the material, 2 in the \emph{few-source} arm and 7 in the \emph{many-source} arm.

\subsection{Results}

The substrate validates. A trained cartridge beats its own initialization by +10.50 and +10.80 points at the two seeds, consistent with the published recipe. No memory budget in our sweep met the accuracy bar and a 10$\times$ compression bar together. The 10$\times$ bar was set in the registered record from the records' lengths, before any training run. That the published LongHealth-specific savings figure is also up to 10$\times$ \citep{eyuboglu2025cartridges}, while the headline 38.6$\times$ is a mean across that paper's benchmarks, is a coincidence the record notes as such, so the bar cannot be mistaken for one tuned to the published result (Appendix~\ref{app:setup}). Prompt compression, shortening or summarizing the text itself rather than training the cache \citep[e.g.,][]{jiang2023llmlingua}, is the natural rival at these ratios, and there may be stronger methods than the recipe we tested on either path, but the published comparison finds compression baselines degrading at ratios as low as 2$\times$ \citep{eyuboglu2025cartridges}. The initialization itself is a strong baseline rather than an empty one. The recipe initializes a cartridge of budget $p$ from the prefilled cache of the first $p$ tokens of the record's own text ($p$ swept from 128 to 8{,}192), and that untrained starting point already scores 0.487 against a held-out floor of 0.415. Training adds the further +10.5 points to 0.592, and it also displaces. Over the 1{,}000 evaluated answers, the trained cartridge loses 90 that its initialization had right while gaining 195.

\begin{figure}[h]
\begin{center}
\includegraphics[width=0.85\linewidth]{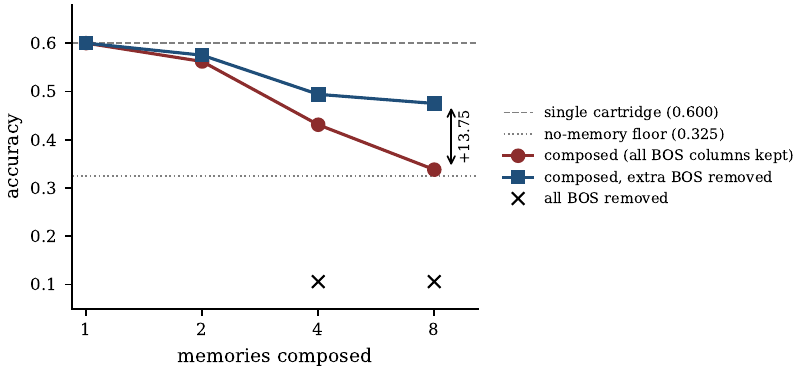}
\end{center}
\caption{Accuracy under composition. The item-matched single-cartridge baseline holds 0.600. The composed memory falls with the number of memories. The same memory with extra BOS columns removed recovers about half the fall, and removing every BOS column collapses the memory to 0.106, below the no-memory floor. Removing an ordinary column instead of a BOS column (not shown) changes nothing.}
\label{fig:compose}
\end{figure}

Composition degrades accuracy, and about half of the loss is attributable to the presence of the extra BOS columns (Table~\ref{tab:compose}, Figure~\ref{fig:compose}). At 8 memories, the raw collapse is 26.25 points, and removing the extra BOS columns recovers 13.75 of them ($p = 4.7 \times 10^{-4}$), leaving a net penalty of 12.50 points, and there is a net penalty of 10.625 at 4 memories. The control's verdict label is exactly one item wide: recovery of 13.75 points clears the half-the-collapse threshold of 13.125 by 0.625 points, and one item on this 160-item set is worth 0.625 points. The significance survives every single-item perturbation. The two extra checks show that one start marker is required, and extra ones are harmful: Deleting an ordinary column changes nothing at any composition size, while deleting every BOS column destroys the memory outright, with accuracy 0.106 and the same option chosen on 160 of 160 items.

\begin{table}[h]
\caption{Composition penalty against the item-matched single-cartridge baseline (each item scored against the cartridge trained on its own source record), before and after the start-marker control. Negative is worse.}
\label{tab:compose}
\begin{center}
\begin{tabular}{lcccc}
\toprule
Memories composed & Raw penalty & $p$ (raw) & Net of start markers & $p$ (net) \\
\midrule
2 & $-3.750$ & 0.362 & $-2.500$ & 0.572 \\
4 & $-16.875$ & $2.5 \times 10^{-5}$ & $-10.625$ & 0.0060 \\
8 & $-26.250$ & $3.3 \times 10^{-9}$ & $-12.500$ & 0.0055 \\
\bottomrule
\end{tabular}
\end{center}
\end{table}

Co-visible training did not protect. At 8 composed memories the penalty is $-22.5$ points in the few-source arm and $-22.5$ in the many-source arm, identical to the decimal, and the penalty grows with the number of memories in both arms. The source-count question, however, is left open by the data itself. Both co-visible arms end close enough to the no-memory floor at 8 memories (0.375 against 0.325) that neither is measurably better than no memory at all, and a comparison between two memories that are both at the floor cannot say anything about the number of sources. The failure is not one of sensitivity, since the same design detects the no-mixing collapse of 26.25 points at $p = 3.3 \times 10^{-9}$ and the arms are balanced with budgets identical to the token. The composed memories are simply too degraded for the contrast to bite (the guard that formalizes this, and the per-size arm penalties, are in Appendix~\ref{app:setup}).

\subsection{Analysis}

About half of the 8-memory collapse is an artifact of how composed caches are assembled, and about half survives the repair we tested. The artifact half has a mechanical fix, removing the duplicated start markers, and the surviving half resisted the one repair we tried, 30\% co-visible training. Repairs we did not test could still reduce it, among them position-layout choices, other special-token handling, calibration or selective recomputation after composition \citep{yao2025cacheblend}, and richer co-visibility schedules, the last being the published repair of \citet{hardalov2026cas} in their setting.

The surviving half has a location. The LongHealth clinical questions each concern a single patient, so composition never removes links a question needs, and the loss here is interference from co-loaded material and probably not a missing-relationship failure. On the published disagreement, the two results need not conflict, since the composability demonstration composed two cartridges and the collapse appeared at twenty. We note as well that the demonstration's multi-document questions appear to ask for parallel facts, one from each source, rather than a chained hop, so composition over questions that need links across sources is untested there \citep{eyuboglu2025cartridges}. Our sweep spans part of that gap and finds the penalty growing from not distinguishable at 2 to 12.50 net points at 8. Our result is consistent with the reported collapse and silent on the reported repair, since four axes separate the settings: 8 versus 20 memories, Llama-3.1-8B versus Qwen3-8B \citep{yang2025qwen3}, 2{,}048-token cartridges versus roughly 585-token ones at much higher compression, and the role each source plays in its collection. Near their mixing fraction, on our model and sizes, co-visible training bought no measurable protection. For practical implementation of cartridges, if composed cartridges are unavoidable, our results support stripping the duplicated start-of-text columns down to one to recover about half the loss for free (Section~\ref{sec:discussion}).

\section{Build time: retraining as sources accumulate}
\label{sec:warmstart}

\subsection{Motivation}

This section takes the question: Can rebuilds cost less than building from scratch? Sources accumulate over a memory's life, as when a clinic's memory that serves one patient record this month must serve two the next. This section is about cartridges, and it studies the accumulation architecture that Section~\ref{sec:compose} did not. Instead of keeping one memory per source and co-loading them, one evolving memory is maintained by further training as sources arrive. One alternative to re-training each time new information arrives is a \emph{warm start}, initializing generation $g$'s training from the trained cartridge of generation $g{-}1$ and then training on the new source's data, optionally mixing back a fraction of the earlier sources' training data, called \emph{replay} \citep{rolnick2019experience}, the standard guard against catastrophic forgetting \citep{french1999catastrophic}. We investigate how many warm-started rebuilds a memory survives before its quality falls measurably behind a fresh retrain on the same material, and what the surviving ones cost. We sweep the replay fraction rather than fixing it, because in the nearest published precedent, a warm-start retraining study in image classification, the stage-10 gap to a fresh retrain on the accumulated corpus reads from its tables as negligible when all earlier data is replayed and about 17 points when none is \citep{shen2024warmstart}.

\subsection{Setup}

The chain starts from a base cartridge trained on one clinical record and adds one new record per generation. Each generation's cartridge warm-starts from the immediately previous generation's cartridge only, forming a ``chain'' over time, and runs through 13 generations at replay fractions 0, 0.5, and 1, with every cartridge at the 2{,}048-token budget. At replay fraction $r$, the rebuild's training data mixes back a share $r$ of the earlier sources' self-study examples alongside the new source's. This is distinct from Section~\ref{sec:compose}'s co-visible mixing fraction, which concerns what other sources a cartridge sees while being trained in the context of co-loading. Training material is self-study conversation over the records (Section~\ref{sec:compose}) and never includes the evaluation items.

Fresh full retrains on the same corpus state provide a moving ceiling, recomputed at generations 1, 2, 4, 8, and 13. A fresh retrain differs from the chain in both accumulated history and initialization, and cartridge-specific initialization on its own is known to improve training \citep{hardalov2026cas}, so a second control retrains from scratch using the chain's own initialization, leaving accumulated history as the only difference. A stale floor, the base cartridge never updated, anchors the other end. Evaluation covers 280 items, 20 per record, over 50 trained cartridges and 14{,}000 measurements. The item set is fixed across generations while the corpus grows, so every curve, ceilings included, rises mechanically as the memory comes to cover more of the items, and the quantity of interest is always a chain's gap to its same-generation retrain rather than any absolute slope. At this corpus size only a 10-point gap is reliably detectable (the power arithmetic is in Appendix~\ref{app:setup}).

\subsection{Results}

The registered outcome is a lower bound. For the replayed chains, the gap to the fresh retrain never reaches 10 points, the smallest gap this corpus can detect, through generation 13, so their tolerable chain depth can only be stated as more than 13. The no-replay chain is the one exception, below.

\begin{figure}[h]
\begin{center}
\includegraphics[width=0.55\linewidth]{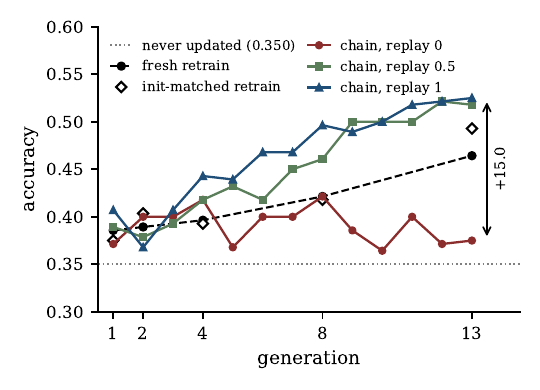}
\end{center}
\caption{Chain accuracy by generation at three replay fractions, against the fresh retrain and the initialization-matched retrain recomputed at generations 1, 2, 4, 8, and 13, with the never-updated base as the floor. The chains cannot be told apart through generation 4 and end 15.0 points apart at generation 13, and the no-replay chain ends level with never updating. Nearly every chain-to-ceiling gap is smaller than its cell can distinguish, and the one exception is the no-replay chain at generation 13.}
\label{fig:warmstart}
\end{figure}

What separates outcomes is replay (Figure~\ref{fig:warmstart}). At generation 13, the full-replay chain sits 15.00 points above the no-replay chain, 0.5250 against 0.3750 at $p = 6.5 \times 10^{-6}$, the largest effect in the experiment, and the two chains cannot be told apart through generation 4. One decay survives every check. The no-replay chain at generation 13 falls 8.93 points below the fresh retrain ($p = 0.0031$) and 11.79 points below the initialization-matched retrain ($p = 0.0001$), so the decay is significant against both, survives the control, and grows, which pins the loss on accumulated history and not on the warm start's starting point.

Against the stale floor of 0.3500, thirteen successive warm-start updates without replay buy $+2.50$ points at $p = 0.51$, nothing measurable over never updating at all, while the same updates with half or full replay are worth $+16.79$ and $+17.50$ and a fresh retrain $+11.43$. The chain's shape matches neither of the two shapes admitted in advance, monotone decay, the shape of a published precedent for incrementally maintained search indexes \citep{singh2021freshdiskann}, or an early drop that then plateaus. The no-replay chain runs level with its ceiling until a late drop, and the full-replay chain rises (Appendix~\ref{app:setup}). One oddity deserves note. The full-replay chain ends 6.07 points above the fresh retrain itself, a gap too small for this corpus to tell from zero. If it is real, the natural cause is cumulative exposure, since by generation 13 the chain has trained over the earlier records at every generation while the fresh retrain sees them once, pointing toward a similar result to one reported by \citet{eyuboglu2025cartridges}, that cartridges become monotonically better with more training.

Replay also sets the cost. At generation 13, one no-replay increment is roughly 15$\times$ cheaper than a fresh retrain on the same corpus state, and it is the arm that decays. One full-replay increment costs about 6\% more than the fresh retrain. And one half-replay increment costs about half of a fresh retrain while ending within a point of full replay, 0.5179 against 0.5250, a difference too small for this corpus to tell apart. So the cheapest rebuild that preserved quality here cost about half a from-scratch retrain, and where between 0 and 0.5 that stops being true is unmeasured.

\subsection{Analysis}

Asked how deep a warm-start chain can go, the data answer more than 13 at replay 0.5 and 1. At replay 0 the chain falls measurably behind both retrains by generation 13, so no depth claim is made for it.

The no-replay chain reads as the co-preparation requirement of Appendix~\ref{app:threshold} arriving at training time. Each generation trains without the earlier records in view, the accumulated material decays until thirteen updates are worth nothing over none, and the initialization-matched control pins the loss on that history. Price scales with the replay fraction, but quality does not fall in step, since half replay preserved what full replay preserved.

Two scope notes close this experiment. The published answer to accumulating sources is one cartridge per document \citep{hardalov2026cas}. Between this section and Section~\ref{sec:compose}, the paper prices both accumulation paths on one corpus. The one-memory-per-source path pays the composition penalty of Table~\ref{tab:compose}, and the one-evolving-memory path pays for replay. A third path, a bounded working set of mutually conditioned cache chunks maintained in place, appears only as a precursor measurement (Appendix~\ref{app:precursor}). Both prices rest on one warm-start rule, one corpus, one cartridge size, and one seed. For practical implementation, rebuilds should be taken warm only with budget for replay, since in these chains the replay fraction at 0.0 is cheap but buys nothing, while rebuilds at half to all of a fresh retrain's cost preserve quality (Section~\ref{sec:discussion}).

\section{Query time: serving corrections beside a built memory}
\label{sec:corrections}

\subsection{Motivation}

This section takes the question: Can corrections be served without rebuilding at all? It is the introduction's query-time option. The memory about Bob's finances is kept as built in January, and the March news arrives beside it instead of adding to it. Memories go stale because facts change, and the practical question is whether the change can be delivered to a built memory cheaply, as an interim measure between rebuilds. The stale material reaches the model in one of two ways, re-read as text or served as its saved prefilled cache, and updates are delivered either as pasted text or as key-value spans written into that cache.

\subsection{Setup}

Each item is a 6{,}000-token base text in which one fact is out of date. The text still states an older version, and the current version arrives only in the accompanying updates. Alongside the base text comes a pile of single-sentence update messages, exactly one of which corrects the queried fact. The rest are filler updates that revise other facts of the same kind, and every pile is brought up to its swept size, from 1 to 512, by adding fillers. Revision sentences are realistic and wrapped in the same prose the base text uses. The pile is presented all at once at each size, and policies that maintain a memory incrementally as updates stream in are outside this experiment's scope. Questions in the primary experiment are two-hop in the near-identical regime (the harder of the two distractor levels, where the wrong options are facts of the same type about confusable entities), and a tier sweep and a robustness grid vary tier, regime, and format.

Two references bracket every policy, and nine policies compete, six serving the pile as text and three writing updates into the stale text's saved cache (Table~\ref{tab:policies}). A \emph{padding} condition separates what the accompanying sentences say from how much there is to read, in which the $k-1$ fillers are replaced by sentences that revise nothing, drawn from other texts' filler paragraphs in the same register (Appendix~\ref{app:setup}). It runs at pile sizes 32 and 512, and it is a control rather than a tenth policy, since it removes the competing revisions that a real pile contains.

\begin{table}[h]
\caption{The two references, the nine policies, and the control. No policy may touch the generator's hidden supersession labels, because arrival order is the only recency signal a deployed system has.}
\label{tab:policies}
\begin{center}
\begin{tabular}{llp{3.5in}}
\toprule
 & Name & What it does \\
\midrule
References & rebuild & reads the corrected text in full, the full-context reference \\
 & stale & reads the outdated text alone, the floor \\
\midrule
Text policies & paste everything & the whole pile pasted beside the stale text \\
 & paste newest-first & the pile ordered newest first \\
 & paste by relevance & the pile ordered by relevance to the question \\
 & paste de-duplicated & near-duplicates removed at a tuned similarity threshold \\
 & paste merged & only the newest update per fact kept, keyed by arrival order alone \\
 & RAG (tuned) & only the top-ranked updates placed in context (Section~\ref{sec:rag}) \\
\midrule
Injection policies & independent & each update captured alone, written into the saved cache \\
 & conditioned & each update captured with the base text in view, then written in \\
 & merged & conditioned capture, only the newest update per fact served \\
\midrule
Control & padding & the $k-1$ fillers replaced by token-matched sentences that revise nothing \\
\bottomrule
\end{tabular}
\end{center}
\end{table}

Items pass two screens before they count. Reading the corrected text alone (nothing preceding), the model must answer correctly, since otherwise no update could be expected to help. And reading the stale text alone, it must answer incorrectly (using the stale text), since otherwise there is nothing to rescue. Of 328 fully evaluated items, all 328 fail from the stale text and 228 pass the corrected read, so the screen only removed items whose corrected text the model could not read. Failing the stale read usually means giving the outdated value, but on 30 items the model picks some other wrong option instead, so contrasts that depend on the outdated value itself are also reported on the 198 items whose stale read produced it (full screening arithmetic in Appendix~\ref{app:setup}).

\subsection{Results}

\paragraph{No policy is workable at any pile size.} We call a policy workable if it ends within 10 points of the full-context reference. The tolerance is a registered number whose use as this bar is a convention adopted during analysis (Appendix~\ref{app:setup}). The reference reads the corrected text alone, so on the screened scale it is 1.000 by construction rather than re-measured at each pile size. Whether a full-context read would itself degrade with the update pile appended is untested here, so the quoted gaps are gaps to an idealized ceiling rather than to a pile-carrying one. On the two-hop items the best single cell any policy reaches is 0.357, and the best policy pooled over pile sizes, tuned retrieval, reaches 0.287, so every policy at every pile size ends at least 64.3 points short of the reference, more than six times the workable line. The closest approach anywhere in the program is the single-hop sweep below, at its plateau, and it is still 20.1 points short. This suggests that under the conditions of this particular experiment, there is no staleness level to schedule rebuilds around. Table~\ref{tab:collapse} gives rescue rates at the smallest and largest pile sizes, reported on the tier sweep's own two-hop screen rather than the primary screen's (Appendix~\ref{app:setup}).

\begin{table}[h]
\caption{Rescue rates by policy at the smallest and largest pile sizes, screened scale (the tier sweep's two-hop screen, $n = 255$, with ceiling 1.000 and floor 0.000 by construction). At pile size 1 there is nothing to order, prune, or rank, so all six text policies coincide. The best pile-512 value is bold.}
\label{tab:collapse}
\begin{center}
\begin{tabular}{lcc}
\toprule
Policy & pile 1 & pile 512 \\
\midrule
Paste everything / de-duplicated & 0.341 & 0.090 \\
Paste newest-first & 0.341 & 0.078 \\
Paste by relevance & 0.341 & 0.196 \\
Paste merged & 0.341 & 0.220 \\
RAG (tuned) & 0.341 & \textbf{0.235} \\
Inject, conditioned & 0.337 & 0.016 \\
Inject, merged & 0.337 & 0.067 \\
Inject, independent & 0.031 & 0.035 \\
\bottomrule
\end{tabular}
\end{center}
\end{table}

\paragraph{A lone correction is used on a third of items.} With a single, relevant, unambiguous correction beside the stale text, every text policy and the conditioned and merged injection policies rescue about a third of guaranteed failures, roughly 0.34, and the outdated account wins most of the rest. The delivery mechanism doesn't change the result, since pasted text and conditioned injection sit within half a point of each other. Only independent injection, the update captured without the base text in view, is used on almost no items, 0.031.

\begin{figure}[h]
\begin{center}
\includegraphics[width=0.55\linewidth]{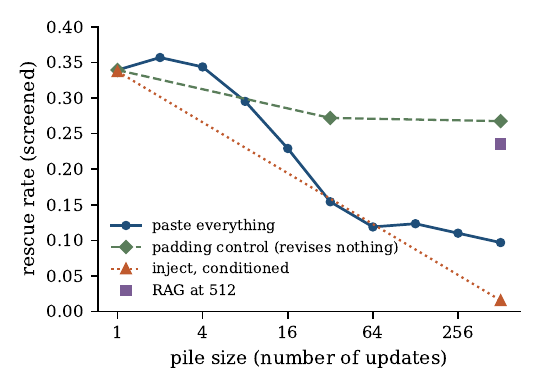}
\end{center}
\caption{Rescue rate against pile size on a log axis. Pasting everything falls as real revisions accumulate, and the same pile sizes padded with token-matched sentences that revise nothing fall far less, so most of the fall is caused by what the sentences say. The padding series is a control rather than a deployable policy, since it removes the competing revisions a real pile contains, which is why it can sit above every policy at 512. Conditioned injection collapses fastest, and tuned retrieval is the best policy endpoint at 512. The plotted series come from different launches, each under its own screen, so endpoints differ from Table~\ref{tab:collapse} by less than half a point.}
\label{fig:pilesize}
\end{figure}

\paragraph{Use falls as revisions accumulate.} Paste-everything falls from about 0.34 at one update to under a tenth at 512, with most of the decline between pile sizes 8 and 64 (Figure~\ref{fig:pilesize}). Policies that prune, order, or select retain more. At 512, tuned retrieval holds 0.235, paste merged 0.220, and paste by relevance 0.196, against 0.090 for pasting everything. The injection policies retain least, at 0.016 to 0.067. At pile 512 the stale answer is chosen on 64\% of screened items and a third option on 26\%. The padding control confirms the damage comes from what the fillers say rather than their bulk.

\paragraph{Text beats injection, most at scale.} The best text policy beats the best injection policy at 8 of 10 pile sizes, each past that comparison's MDE, with a pooled lead of 9.45 points. Pooled over pile sizes, the best text policy is tuned retrieval at 0.287 and the best injection policy is merged injection at 0.192 (per-size statistics in Appendix~\ref{app:setup}). The two sides are equal at pile size 1 and separate as revisions accumulate, so what injection loses is not the ability to deliver an update but the contest among many.

\begin{figure}[h]
\begin{center}
\includegraphics[width=0.55\linewidth]{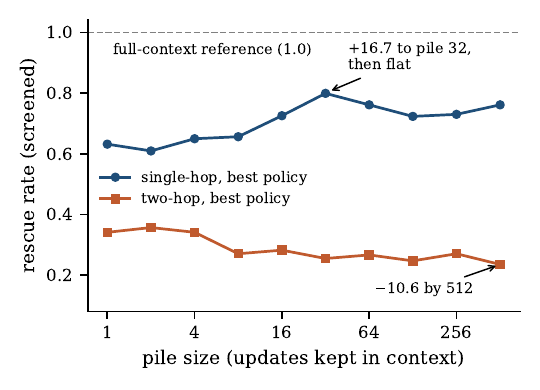}
\end{center}
\caption{Rescue rate against pile size for each question tier's best text policy, near-identical regime, both tiers swept at all ten pile sizes (screened scale, with single-hop $n = 448$ and two-hop $n = 255$). Single-hop rises to a plateau at pile size 32, with 32 against 512 inside its MDE. Two-hop never gains and ends 10.6 points below its pile-1 value. The winning policy differs by tier, an ordered paste against tuned retrieval, so the curves compare each tier to its own baseline rather than tier to tier.}
\label{fig:tiers}
\end{figure}

\paragraph{Updates kept in context help single-hop questions and hurt two-hop ones.} A tier sweep runs both question tiers over all ten pile sizes in the near-identical regime, each scored on its own screen ($n = 448$ single-hop, $n = 255$ two-hop) with its own best text policy, an ordered paste for single-hop and tuned retrieval for two-hop (Figure~\ref{fig:tiers}). The shapes differ. The single-hop curve rises 16.7 points from pile size 1 to 32 ($p = 1.2 \times 10^{-10}$, MDE 7.4) and then stops, with the change from 32 to 512 inside its MDE. The two-hop curve never gains, with pile size 2 against 1 inside its MDE, and ends 10.6 points below its pile-1 value at 512 ($p = 2.0 \times 10^{-4}$, MDE 7.8). For a single-hop question answered from an update stream, keeping about 32 updates in context is worth 16.7 points over keeping one, and keeping more buys nothing measurable, while for a two-hop question, more updates are never worth anything. Even at its plateau the single-hop cell sits 20.1 points short of the reference, past the workable line, so this is an operating point rather than a rebuild schedule. Two cautions travel with it. The winning policy differs by tier, so the contrast is each tier's shape against its own baseline rather than a clean tier effect, and the padding condition ran only on the two-hop cell, so whether the single-hop gain comes from the updates' content or their bulk is untested.

\paragraph{Robustness.} The tier shapes and the text-over-injection lead reproduce across seven held-out combinations of question tier, distractor regime, and answer format, and the distance to the full-context reference exceeds the workable line in every group. Screening pass rates differ across the groups, so no cross-group differencing is done. The full grid, 91{,}350 measurements, is in Appendix~\ref{app:setup}. A companion experiment on repeated revisions to one fact, the single case where the method under test holds a structural advantage (only a merged memory encodes recency by construction), also comes back against it. The recency-blind relevance policy pays only about 1.6 points of extra decay per revision, and merged injection never beats the same merge run as text (Appendix~\ref{app:revisions}).

\begin{figure}[h]
\begin{center}
\includegraphics[width=0.55\linewidth]{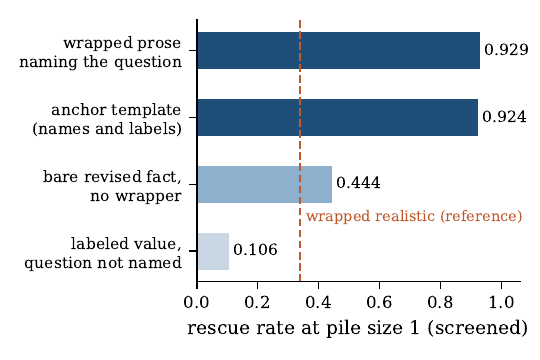}
\end{center}
\caption{Rescue rate by update form at pile size 1 (pasted, though conditioned injection matches within a point), on the 198 items whose stale read produced the outdated value. The dashed line is the wrapped realistic reference on the full screened set. Naming the question moves an update from partial use to near-full use, and a labeled value that names nothing is the weakest form.}
\label{fig:binding}
\end{figure}

\paragraph{What makes an update bind.} What is it about an update's form that decides use? On the 228 screened items at pile size 1, a factorial over update forms separates the features a realistic revision carries together in natural text: the prose wrapper, an explicit label on the new value, and naming the question the update answers. Five forms carry the identical correction through both mechanisms, pasted text and conditioned injection. Naming the question is the factor (Figure~\ref{fig:binding}). Wrapped prose that names the question is used on 0.929 of items and an anchor template that names and labels at once (``Query on file: \{question\}. Verified current finding: \{answer\}'') on 0.924, against 0.444 for the bare revised fact and 0.106 for a labeled value that names nothing, with the wrapped realistic reference, the revision sentence of the setup above, near 0.34. The pre-registered naming contrast is $+83.3$ points on the full 228-item screen ($+82.3$ on the 198 whose levels are shown; $p = 1.3 \times 10^{-57}$, MDE 16.9), it replicates on a fresh item manifest sharing no texts with the original ($+77.5$ pasted, $+77.8$ injected), and the delivery mechanism changes no cell by more than a point.

\paragraph{Naming works by compliance, not recognition.} A wrong-naming control, registered with its predictions after the factorial's result and before its own run, serves the anchor template carrying the item's own question but a well-typed false value, a decoy drawn from the same relation's other answers. If naming worked because the model recognizes an update's stated relevance, the decoy should not be adopted. It is adopted on 0.877 of items (95\% CI 0.828 to 0.914) against a no-update base rate of 0.009, at $p \approx 5 \times 10^{-60}$, and the true answer is produced exactly never. The decomposition repeats the factorial's. Naming alone moves adoption of a false value by $+82.0$ points, while the template and the delivery mechanism move nothing (mechanics in Appendix~\ref{app:setup}).

\paragraph{Binding holds on every substrate.} A companion experiment on the clinical records serves the anchor form as an addendum to built memories over superseded facts, on every substrate including a trained cartridge. We say an update \emph{binds} when the model's answer follows the update instead of the memory's original content. It binds on every substrate tested, moving accuracy on superseded facts from 0.105 to 0.655 in the primary run with no items lost, and by 45.0 to 64.5 points across substrates and replications. On the conditioned arm a trained cartridge and a quality-matched cache are not distinguishable (per-arm statistics in Appendix~\ref{app:setup}). A descriptive check on a second model, Qwen3-8B, repeats the naming direction at clearly higher levels under that model's own screen (Appendix~\ref{app:setup}).

\subsection{Analysis}

The wall at a single update is a conflict-level property of the model. The paste policies re-read the stale text and the correction together in one fresh prefill, so from a co-preparation standpoint they hold every advantage a memory can have, and they still fail on two-thirds of items, and conditioned injection fails at the same rate.

Accumulation adds a second, separable cost, and the padding contrast places it in the updates' content. A large block of inert prose costs little, while the same volume of other revisions is costly. The policy ranking at scale is consistent with that reading, since every policy that reduces the number of competing revisions in view, by ranking, merging, or selecting, retains more than pasting everything, and tuned retrieval, which serves the fewest, retains most. Injection retains least. Updates captured into the cache lose the contest among many revisions faster than the same updates re-read as text, and this is where the substrates separate after being indistinguishable at a single update. Independent injection fails even alone, which is the co-preparation thesis in miniature, that an update captured without the base text in view never binds to it.

What moves the wall is wording, and the boundary is not where the delivery machinery is. Pasted text and conditioned injection agree within a point in every cell of the factorial, so the pathway into the model is interchangeable, and what decides use is whether the update's own wording names the question it bears on. Why naming works is settled by the control, and the answer is compliance. The model adopts a well-typed false value nearly as readily as the true one, so an update channel behaves as an injection surface (Section~\ref{sec:conflict}), and control of update wording is control of the answer, a susceptibility documented a model-generation ago, when distractor sentences constructed to overlap the question fooled extractive readers the same way \citep{jia2017adversarial}. Naming is also the query-time face of co-preparation. At build time the model uses a cross-fact connection only if the connection was present in the pass that prepared the memory (Appendix~\ref{app:threshold}), and at query time it uses an update only if the update's words state its connection to the question. In both places the connection is the scarce good, and the model neither infers an unstated connection nor verifies a stated one.

This experiment was designed to yield a rebuild schedule, meaning a staleness level at which rebuilding becomes necessary and below which serving updates suffices. None exists in the data. With a gap of at least 64 points at every pile size, there is no level to schedule around. Three cautions scope the wording result. The factorial and its control ran at a single update, and whether a named update keeps binding or a mis-named one keeps hijacking while many revisions accumulate is unmeasured. Deployed updates arrive question-agnostic, so using this lever requires knowing, or generating, the questions an update bears on, and our compliance finding raises the price of generating them wrong (Section~\ref{sec:discussion}). For practical implementation, corrections served beside a stale memory are not a substitute for rebuilding on this model. If updates must be served in the interim, serve few, by ranking, merging, or retrieval, and write each so that it names the question it answers. Some memory systems split memory across mechanisms, with parametric memory in the weights, editable text notes, and search \citep{engram2026study}. Concurrent work reaches staleness from the system side, finding agents act on superseded constraints in about three-quarters of episodes when verification is budgeted \citep{nakayashiki2026stale}, and proposing explicit revocation of contradicted memories as the repair \citep{zhou2026tepa}. What this section measures is the narrower slice any such system faces when new information contradicts an already-built store, whichever mechanism holds it.

\section{Discussion}
\label{sec:discussion}

\subsection{Cache against cartridge}

The two substrates were tested side by side so that a limit could be attributed either to a substrate or to precomputed memory as such, and most of the evidence points at the latter, with one measured exception on each side. At a single update the mechanism is interchangeable. Pasted text and conditioned injection agree within a point across the wording factorial, the anchor addendum binds on every substrate with lifts of 45.0 to 64.5 points, and under matched base quality cache and cartridge are not distinguishable on the conditioned arm, equivalent to within about 10.6 points, the comparison's MDE, in a run that detected 55-point effects on the same items (Appendix~\ref{app:setup}). The cartridge's one measured edge is that it follows independently injected updates more readily than the matched cache, by 12.0 and 14.5 points in the two runs. The cache side's one measured edge appears under accumulation, where updates re-read as text beat updates captured into the cache at 8 of 10 pile sizes with a pooled lead of 9.45 points.

The build-time penalties fall on each side by design rather than by substrate. The piece deficit of 48.5 to 79.5 points belongs to caches assembled from isolated pieces, the composition penalty of 10.6 to 12.5 net points belongs to cartridges, and the order-of-magnitude difference traces to the questions, since the clinical questions never need links across sources. The remaining choice between the substrates is the plain economic one, one prefill against training time, full size against compression, with the reminder that a trained memory's accuracy on the records sits 23.0 to 29.5 points below the full-context read at the sweep's primary budget, and 11.4 points below at its best budget, which compresses only 1.3$\times$ (Appendices~\ref{app:setup} and~\ref{app:measure}).

\subsection{One hop against two}

The tiers, one-hop and two-hop, separate on every measurement. At build time, the two-hop threshold is 0.82 to 0.92 while single-hop thresholds, where measurable, are 0.34 to 0.58 (Appendix~\ref{app:threshold}), and unprepared accuracy starts at 0.195 to 0.510 on two-hop questions against 0.635 to 0.795 on single-hop, because the fact-bearing piece is intact in the composed memory. At query time, best-policy rescue at pile size 1 is 0.21 to 0.34 in the two-hop groups against 0.41 to 0.63 in the single-hop groups, and the shapes diverge, a single-hop rise of 16.7 points to a plateau near 32 updates against a two-hop curve that never gains and ends 10.6 points down. What co-preparation buys is the connection between facts, and questions that need no connection are forgiving. The same split shows in the motivating observation of \citet{eyuboglu2025cartridges}. A cartridge trained by plain next-token prediction memorizes its corpus at 107$\times$ less memory than the full cache yet fails queries that ask for more than recall, so recall of stated facts is the cheap part of precomputed memory, and connection is what costs. The same reading fits a concurrent serving system that composes independently prepared tool-schema caches with no measured accuracy loss, in a workload where no query needs links across sources \citep{fang2026recache}. One reading of the query-time shapes is that an update which settles a question outright can only add evidence, while updates feeding a two-hop chain compete with the connection the question must traverse. The tiers' winning policies differ, which keeps this a reading rather than a result. Our cartridge experiments are not tiered by hops, because the records' questions are not labeled by hops, so that comparison is open here. \citet{eyuboglu2025cartridges} do elicit multi-hop questions in their synthetic diagnostic data, within a single document.

For practical implementation, the split cuts by workload. Lookup-style traffic, one stated fact per question, tolerates partial preparation, staleness, and accumulated updates far better than synthesis-style traffic, so the multi-hop share of a workload is what sets the co-preparation bill and the rebuild cadence. A system that routes multi-hop questions to a full-context read while serving single-hop questions from memory would concentrate the memory's value where it survives.

\subsection{What the results price out}

A saved cache over material that is not changing costs one prefill and repays itself on its first reuse, and nothing here disturbs that. What the results remove is every cheaper variant of staying current: preparation cannot be made partial (Appendix~\ref{app:threshold}), separately prepared parts recombine only at a penalty that grows with their count (Section~\ref{sec:compose} and Appendix~\ref{app:threshold}), a warm rebuild preserves quality only with replay, at half to all of a fresh retrain's cost as measured, possibly less (Section~\ref{sec:warmstart}), and no policy for serving corrections beside a built memory (besides the wording ablation) comes within 20 points of the reference anywhere, or within 64 on the two-hop items (Section~\ref{sec:corrections}). A memory's serviceable life is therefore the interval between changes to the facts it covers, and its value is the prefills saved inside that interval minus its build cost, which for a cartridge includes training. In the one fully metered case, the retraining chains, training cost exceeded the cost of the 14{,}000 evaluations run on the results by a factor of about 600 (Appendix~\ref{app:setup}). We do not quote a numeric break-even, because it depends on a workload's query rate. The rebuild cadence is set by how often facts change, because no workable staleness level exists to stretch it.

For someone building such a system, the results support concrete advice. Prepare a memory in one joint pass over everything it should serve, and treat the full preparation as the price of admission. Do not assemble a memory from pieces prepared separately, at any piece size. If composed cartridges are unavoidable, remove the duplicated start-of-text columns and keep exactly one, which recovers about half the loss for free. Take rebuilds warm only with replay, and budget for it. In our chains, full replay cost about as much as a fresh retrain, half replay preserved the same quality at about half that cost, and nothing below half is measured. Rebuild when facts change rather than relying on corrections served beside a stale memory. If updates must be served in the interim, reduce the number of competing revisions in view, by retrieval or by merging down to the newest version per fact, and keep no more than approximately a few dozen updates in context, since past roughly 32, additional updates bought nothing on single-hop questions and cost accuracy on two-hop ones in our tests on our configuration. Above all, write each update so that it names the question it answers, which raised use from a third to over nine-tenths in our tests and replicated, and guard whatever writes those names, since a false update that named the question was followed nearly as readily. And measure the evaluator before trusting a number, since format, template provenance, and item selection each move results by more than many published gaps (Appendix~\ref{app:measure}).

\subsection{Ways forward}

What follows is informed speculation, separated from the measurements above. Five measurements are one run away, each converting a named unknown into a number. The padding control on the single-hop cell of the tier sweep would settle whether that tier's gain is content or bulk. The wording factorial swept over pile sizes would show whether a named update stays bound as revisions accumulate. Replay fractions between 0 and 0.5 are the most grounded cost lever this paper's own data suggest. Composition at 16 to 20 cartridges would meet the published operating point of \citet{hardalov2026cas}. The fifth is the visibility sweep at 48{,}000 tokens and beyond (Appendix~\ref{app:threshold}).

The naming result points at a design. Updates arrive question-agnostic, so exploiting the lever means generating, at update arrival and with the same model, the questions an update bears on, and storing the update annotated with them so the annotations do the naming at serving time. Because the wrong-naming control showed that naming hijacks regardless of content, a wrongly generated name is actively harmful, and the design needs precision. Concurrent work already optimizes edit wording directly, learning prompt constructions that make in-context edits reliable and specific \citep{wang2026moike}. The two-thirds of items where a lone realistic correction loses to a coherent stale text looks like a training target, since a model could be taught to prefer a dated revision over the account it contradicts, which no serving policy can force from the outside. A concurrent step in that direction trains a model to re-attend to cached rules when later steps supersede them \citep{yu2026rules}. The forming cost has untested dials as well: fewer self-study conversations per source, replaying distilled summaries instead of full conversations, the previous cartridge as a distillation teacher, and seed prompts tailored to a specialized workload rather than the recipe's deliberately generic five \citep{eyuboglu2025cartridges}. Serving systems already reuse and route precomputed state, through modular prompt-level key-value reuse \citep{gim2024promptcache}, paged prefix caching \citep{kwon2023paged}, cached-chunk fusion for retrieval \citep{yao2025cacheblend}, and GPU-native injection of independently precomputed fact memories \citep{li2026inferscale}. None studies the update question, and the last is close to the pathway our independent-injection arm prices, though its memories are captured with a small window of preceding context, between our independent and conditioned arms. Finally, architectures with explicit memory or recurrent state change what precomputed memory even is, and the limits measured here are limits of key-value cache memory on one transformer.

\section{Limitations and scope}
\label{sec:limits}

Two limitations dominate. First, every verdict comes from one model, Llama-3.1-8B-Instruct. A systematic second-model comparison was outside this study's scope, and the descriptive check in Section~\ref{sec:corrections} repeats the direction on Qwen3-8B at clearly higher levels, with 141 items and a screen that overlaps the primary one on only 119, so it yields estimates rather than a model comparison. The single-update conflict residue in particular is a behavioral property that models trained differently on contradiction could place elsewhere. Second, the tested lengths stop at 24{,}000 tokens (Appendix~\ref{app:threshold}). The deployments with the most to gain from precomputed memory are exactly those with much longer material, where a constant fraction is most expensive if it holds and the approach is most attractive if it bends, and the slope's confidence interval constrains nothing outside the tested range.

Most quantities here are bounded by their tested ranges and constrain nothing beyond them: 13 generations, 24{,}000 tokens, 512 updates, revision depth 5, and pile size 1 for the wording factorial and its control. The warm-start answer is a lower bound resting on one warm-start rule, one corpus, one cartridge size, and one seed, and a precedent from incremental index maintenance shows the maintenance rule itself can arrest such decay \citep{singh2021freshdiskann}. In particular, this paper cannot claim anything about language models generally; a threshold compared across cache and cartridge, which was never measured; behavior beyond 24{,}000 tokens; a rebuild schedule, since the gap to the reference leaves nothing to schedule around; the size of the gap to a reference that also carries the update pile, since the full-context reference is measured on the corrected text alone; whether a named update stays bound as revisions accumulate; a clean tier effect on the pile-size shape, since the tiers' winning policies differ and the padding control ran only on the two-hop cell; a numeric amortization break-even, which needs a workload's query rate; or anything about real conversations, which none of these experiments measures. Beyond those, the threshold, staleness, and confusability experiments run on synthetic texts, and LongHealth is the one real-record setting. Rescue rates are on a screened scale and are not comparable to unscreened accuracies in other studies. The start-marker control's verdict label is one item wide, though its significance is not. Single-hop thresholds are directional. The trained ceiling of 0.651 is a property of one recipe at one budget.

\subsubsection*{Broader Impact Statement}

The negative results here bear on deployment. Systems that serve precomputed memory without rebuild discipline repeated stale facts on most items in our tests even when corrections were present in context, a concrete reliability risk in settings such as clinical or legal assistance. The one large, replicated lever we found, wording updates to name the questions they answer, is an implementation choice rather than a model change, and the same lever is a hazard since an update that named a question was adopted nearly as readily when its value was false.

\bibliography{main}
\bibliographystyle{tmlr}

\appendix

\section{Common setup in full}
\label{app:setup}

\subsection{Model, decoding, and hardware}

Every number in this paper comes from one model, Llama-3.1-8B-Instruct, loaded in bf16 from a mirror verified bit-identical to the gated original across all weight shards, except where a second model is named and pinned in Section~\ref{sec:corrections} and in the precursor measurements of Appendix~\ref{app:precursor}, which carry no evidential weight. Decoding is fully deterministic. Free-form answers are generated greedily, and multiple-choice answers are scored by length-normalized mean token log-probability over the options under teacher forcing, a variant of the multiple-choice scoring in the standard few-shot evaluation harness \citep{gao2023harness}, so replication throughout this paper means fresh items under a fresh seed. The chat template is the canonical Llama-3.1 template, vendored and hash-checked at load, because the mirror ships a wrong, much shorter template (Appendix~\ref{app:measure} reports what that error costs), and the template's date field is pinned so the same item renders identically across days. Experiments ran on A100 and RTX A6000 GPUs, whose A100 variants agree bit-for-bit while the A6000 differs in floating-point reduction order, and items are sharded so that all conditions of one item share a device.

Across independent launches, re-measured reference rows in the staleness experiment agree on 385 of 388 items, a cross-launch noise floor of 0.77\% arising from near-tie flips under differing device reduction order, and on 11{,}900 of 11{,}900 in the repeated-revision experiment. No reported verdict rests on a margin below that floor, and no quantity is quoted twice from different launches in the main text.

\subsection{Texts, names, and validation}

The synthetic texts come at five lengths, roughly 1{,}500 to 24{,}000 tokens, a 16$\times$ range, each a sequence of short record-keeping paragraphs about invented people, projects, and places. All names are invented and assembled from inventories of uncommon syllables, the inventories share no three-character prefix, and year ranges for different fact types are disjoint, so that no fact is answerable from pretraining, names of different kinds cannot be confused with each other, and one kind of number can never be misread as another. For two-hop questions, a validation check requires the two hops to sit in different paragraphs, and the answer's paragraph never names the project, so the chain cannot be shortcut. Distractor regimes are three by intent. Near-identical distractors are facts of the same type about confusable entities, same-type distractors concern clearly different entities, and mixed-type distractors are unrelated facts. Appendix~\ref{app:measure} shows the first two are empirically one level of difficulty, so the main text describes regimes as two-level.

The padding sentences of Section~\ref{sec:corrections} are drawn from other texts' filler paragraphs, same generator, same lexicon, same administrative register, token-matched to the real filler updates at each pile size. The no-revision property was audited rather than asserted. On a 3{,}000-sentence sample, none matches the revision pattern and none contains the gold or outdated value.

\subsection{References, floors, and screening in detail}

Two different ceilings appear in this paper. The full-context read exists everywhere and is the reference for every gap we quote. A \emph{trained ceiling}, the accuracy our sweep's primary-budget cartridge reaches, exists only on the clinical records, because that is the only material any cartridge was trained on. No cartridge was ever trained on the synthetic texts, so none of the synthetic gaps can be restated against one. Where both ceilings exist, the full-context read scores 23.0 to 29.5 points above the trained ceiling, the two endpoints being one cartridge measured on two item sets, 0.780 against 0.550 on the full 200-item set and 0.946 against 0.651 on the 149-item independently selected set (Appendix~\ref{app:measure}). The right use of that correction is qualitative. Gaps quoted against full context, here and in the literature, are generous by roughly that much, and we do not subtract it from numbers measured on other material. A floor is computed separately per condition, which is why it is a range rather than one number. On the synthetic texts the measured multiple-choice floor is 0.209 to 0.247 against a nominal 0.25, below chance because items the model could answer without the text are screened out before any experiment runs, and the clinical evaluations carry their own, separately measured floors (0.325 and 0.415 in Section~\ref{sec:compose}).

The correction experiments add the second screen described in Section~\ref{sec:setup}. Two facts keep that screen from driving the results. The floor is 0.000 on the unscreened set as well, because every one of the 328 evaluated items produced a wrong answer from the stale text, so the screen only removed the 100 items whose corrected text the model could not read. And the screen's failure criterion is a wrong answer, not specifically the outdated answer. On 30 of the 228 screened items the stale read chose a distractor rather than the outdated value, which is why contrasts that depend on the outdated value are additionally reported on the 198-item subpopulation where the stale read produced it.

\subsection{Pre-registration and reporting rules, with provenance}
\label{app:prereg}

The program was pre-registered. An append-only record fixed analysis rules before the corresponding data collection, and results are identified by content hashes over item manifests and source, which survive repository history changes. Registered before the corresponding runs: the 200-item minimum behind reported comparisons; replication on fresh items before reporting a result in an experiment's registered favorable direction (thesis-favorable in most experiments, method-favorable where the method under test held the structural advantage); tuning for rival methods; publication of negative results; Section~\ref{sec:warmstart}'s two admissible chain shapes; Section~\ref{sec:compose}'s above-floor precondition for the co-visible contrast; Section~\ref{sec:compose}'s 10$\times$ compression bar, set from the records' lengths before any training run, its coincidence with the published LongHealth-specific figure noted in the record; Appendix~\ref{app:revisions}'s ordering and merged-memory rules; and staleness tolerances of 5 and 10 accuracy points, registered for a planned staleness-tolerance experiment that was not run. Registered by pre-run amendment: the degenerate-configuration bar of Appendix~\ref{app:threshold}, raised from 0.10 to a total climb under 25 points by Amendment 3.3 before the threshold experiment ran; and the wrong-naming control (Amendment 94), whose design, point predictions for both readings, and decision rule were committed before any row existed, and whose compliance outcome is the direction unfavorable to serving-time correction, so it reports on first observation with no replication owed. Adopted during analysis: the use of the registered 10-point tolerance as Section~\ref{sec:corrections}'s workable line, labeled as a convention where it is used. The registered 5-point companion tolerance is not used in the main text.

The minimum detectable effect convention is $\alpha = 0.05$ two-sided at 80\% power for an exact McNemar test \citep{mcnemar1947} on a comparison's discordant pairs, computed per cell from the realized discordance following \citet{connor1987sample}. At 25\% discordance this works out to a 9.85-point MDE, which is why 10 points recurs as the round detectability figure. An observed effect smaller than the MDE can still return a small $p$-value, because the MDE is a statement about power rather than a significance cutoff, and cells where that occurs are adjudicated individually in Appendix~\ref{app:relocated}. The bootstrap intervals of the threshold experiment are confidence intervals on the fits of Appendix~\ref{app:threshold}, a different quantity, and the two are never interchanged.

\subsection{Details relocated from the experimental sections}
\label{app:relocated}

\paragraph{Section~\ref{sec:compose}.} The co-visible arm penalties are $-0.6$ (few-source) and $-5.0$ (many-source) at 2 memories, neither significant, $-12.5$ in both arms at 4, and $-22.5$ in both arms at 8. The registered precondition required at least one arm's composed memory to sit measurably above the floor. Both end within their MDE of it, so the precondition failed and no conclusion about source count is drawn.

\paragraph{Section~\ref{sec:warmstart}.} Power arithmetic: at this corpus size, distinguishing a 5-point gap would need 951 items and a 2-point gap 5{,}944, both recorded as out of reach rather than estimated. Per-cell adjudications inside their MDEs: at generation 8 the full-against-no-replay difference of $+7.50$ points is significant at $p = 0.026$ but inside its MDE of 8.97. Two full-replay cells sit above their moving ceiling at $p < 0.05$, $+7.50$ at generation 8 and $+6.07$ at 13, both inside their MDEs. The generation-13 replay contrast carries 95\% CI $+8.52$ to $+20.41$ against MDE 9.24. The no-replay chain's difference from its ceiling runs $-1.4$, $+1.1$, $+2.1$, $0.0$, $-8.9$ points at generations 1, 2, 4, 8, and 13, with the generation-13 decays at $p = 0.0031$ against the fresh retrain and $p = 0.0001$ against the initialization-matched retrain. Metered cost, per increment at generation 13: 0.34 GPU-hours at replay 0, 2.62 at replay 0.5, 5.35 at replay 1, against 5.03 for a full fresh retrain on the same corpus state. Training cost across the experiment totals 88.78 GPU-hours against 0.147 GPU-hours for its 14{,}000 evaluations. The registered comparison of this experiment against the query-time staleness results, fixed at a 5-point gap, sits below this corpus's measurable range and is recorded as unresolvable rather than rerun at a friendlier level.

\paragraph{Section~\ref{sec:corrections}, policies and grid.} Screening arithmetic: 328 items fully evaluated, all 328 fail the stale read, 228 pass the corrected-text screen, 30 of the 228 fail the stale read on a distractor rather than the outdated value, leaving 198 whose stale read produced it. In the text-over-injection comparison, best text beats best injection at 8 of 10 pile sizes, pooled lead 9.45 points, MDE 2.26, $p = 4.0 \times 10^{-33}$. The robustness grid comprises 91{,}350 measurements over seven held-out combinations of question tier, distractor regime, and answer format at pile sizes 1, 32, and 512. The distance to the full-context reference exceeds the workable line in every group, at 36.8 points or more at pile size 1. The change from pile size 1 to 512 in each group's best text policy is positive in the four single-hop groups, by 10.3 to 13.8 points, inside its MDE in two of the three two-hop groups, and a significant fall of 13.7 points in the third, endpoint differences consistent with the tier sweep's two shapes. Sampled at three pile sizes, they cannot fix a shape on their own. The text advantage over injection appears in the held-out groups as well. Screening pass rates vary from 0.545 to 1.000 across groups, so cross-group comparisons are between different item populations and are stated with both rates.

\paragraph{Section~\ref{sec:corrections}, wording factorial and companions.} The five update forms: \emph{wrapped realistic}, the revision sentence of the setup, the reference; \emph{unwrapped}, the bare revised fact with no wrapper; \emph{prose-named}, a wrapped sentence that explicitly names the question it answers; \emph{labeled-unnamed}, the new value in a labeled field without naming the question; and the \emph{anchor} template, which names and labels at once. Two contrasts were fixed in advance, naming (prose-named against labeled-unnamed) and full marking (anchor against unwrapped). Form levels are reported on the 198-item population, and the registered naming contrast is $+83.3$ on the full 228 and $+82.3$ on the 198. The full-marking contrast is $+46.9$ pasted and $+48.2$ injected, replicating at $+50.0$ and $+49.6$, and the replication manifest passes 284 of 400 items. The clinical companion in full: the anchor addendum moves accuracy on superseded facts from 0.105 to 0.655, $+55.0$ points, 110 items gained and none lost, $p = 1.5 \times 10^{-33}$, replicating at $+45.0$ on held-out patients. Binding holds on every substrate tested, trained cartridge and prefilled caches alike, with lifts of $+45.0$ to $+64.5$, all at $p < 10^{-24}$. On the conditioned injection arm a trained cartridge and a quality-matched cache are not distinguishable, at $+2.0$ points with $p = 0.694$ and a replication at $+3.0$ with $p = 0.471$, equivalent to within about 10.6 points, the comparison's MDE, in a run that detected 55-point effects on the same items. On the independent arm the cartridge is distinguishable, in the opposite direction. It follows the injected update more readily than the matched cache by 12.0 and 14.5 points in the two runs, both significant. The second-model check repeats the wrapped-against-anchor contrast on Qwen3-8B, pinned by revision with a vendored, hash-checked template, text arms only, at pile size 1 under that model's own screen, which passes 141 of 328 items. The pasted realistic correction is used on 0.582 of that model's guaranteed failures (Wilson 95\% CI 0.50 to 0.66; \citealp{wilson1927}) and the anchor addendum reaches 0.943 (0.89 to 0.97), estimates rather than verdicts, and the two models' screens overlap on only 119 items, so the columns are never differenced into a model effect. The wrong-naming control runs on all 228 screened items at pile size 1, 1{,}596 measurements over the anchor, prose-named, and labeled-unnamed forms through both delivery mechanisms, with a no-update arm supplying the decoy's base rate. Decoy values are drawn from the same relation's other gold answers, so every decoy is well-typed, and scoring is free-form decoding classified by containment of the decoy, stale, or true value, which is why its rates are adoption rates on their own scale. Adherence to the stale value collapses from 0.693 to 0.101 under the decoy anchor. The primary comparison is a paired exact McNemar test of decoy adoption against the no-update arm, with a realized MDE of 17.2 points.

\section{Build time: how much co-preparation a memory needs}
\label{app:threshold}

This appendix reports the experiment behind the introduction's premise. Can preparation be made partial? If the requirement is a fixed number of tokens, long texts are cheap to prepare relative to their size and the approach scales. If it is a constant fraction, preparation cost grows with the material indefinitely. Prior work on parallel and modular reuse of precomputed context points at the direction. Independently encoded chunks never attend to one another, the resulting drop is documented and analyzed \citep{ratner2023pcw, yang2025ape, zhang2025attentionentropy}, it falls hardest on questions that must reason across chunks \citep{yang2024revisitingpcw}, and the repairs in that literature recompute, re-encode, or recalibrate what the separate passes missed \citep{yao2025cacheblend, ma2025blockattention, hu2025epic}. What is measured here is the magnitude of the requirement and how it scales.

\paragraph{Design.} Every memory in this experiment is assembled from separately prefilled pieces, and the whole text is always present in the memory, and no condition withholds any content. Each text is cut on paragraph boundaries into pieces that cover it end to end, with no overlap and no gap. Each piece is prefilled on its own, preceded by the chat header and by $f \cdot D$ tokens of the text that immediately precede the piece, where $f$ is the \emph{visibility fraction} and $D$ the text length. After the solo prefill, only the piece's own key-value entries are kept (a \emph{column} is the stored entries for one token position). At $f = 1$ the captured columns are identical to the piece's slice of one joint prefill, so the top of the sweep is exactly the ordinary full prefill, and at $f = 0$ every piece was captured blind. Two mechanical facts make the assembly exact. Attention in this model is causal and the header is byte-identical across pieces, so the header's entries come out identical in every pass and the composed memory keeps one copy. And positions enter attention through rotary embeddings \citep{su2021roformer}, which depend only on the difference between two tokens' rotations, so moving a captured piece to its true position is one exact rotation of its stored keys, with no further model pass. Fifteen values of $f$ from 0 to 1 are crossed with the five text lengths, both question tiers, both answer formats, and two piece sizes, so that piece size and text length are not collinear. The threshold $\tau$ is the visibility fraction at which a monotone fit of accuracy against $\log f$ reaches 95\% of the climb from $a(0)$ to $a(1)$, with bootstrap confidence intervals over items. Configurations whose climb is under 25 points, a bar registered by pre-run amendment (Appendix~\ref{app:prereg}), are excluded from the scaling fit. The replication is the entire sweep rerun on fresh texts and questions from the same generator.

\begin{figure}[h]
\begin{center}
\includegraphics[width=0.85\linewidth]{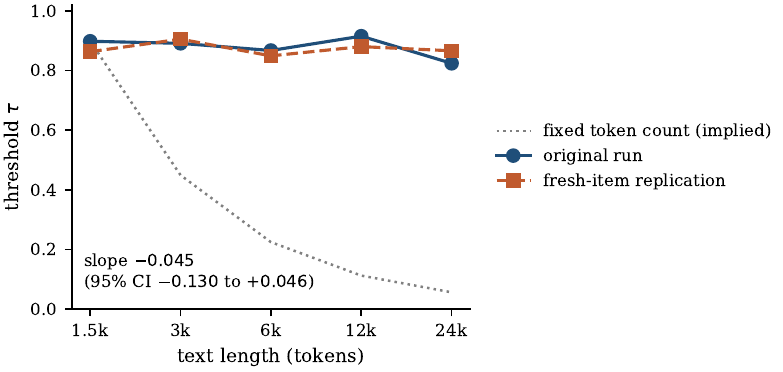}
\end{center}
\caption{The threshold $\tau$ against text length on a log axis, both runs. The dotted curve is the fall a fixed token count would imply, anchored at the shortest length's measured threshold, $0.898 \times 1{,}500 \approx 1{,}350$ tokens. The fitted slope of the measured thresholds is $-0.045$ (95\% CI $-0.130$ to $+0.046$), consistent with a constant fraction over the tested range.}
\label{fig:tau}
\end{figure}

\paragraph{Results.} On two-hop questions the threshold sits between 0.82 and 0.92 at every text length, in both runs (Table~\ref{tab:tau}, Figure~\ref{fig:tau}). The slope of $\tau$ against log length is $-0.045$ (95\% CI $-0.130$ to $+0.046$; replication $-0.046$, CI $-0.128$ to $+0.015$), inside the band fixed in advance for a constant fraction and far from the band for a fixed token count. The same experiment contains a second result. Pieces captured in isolation and then composed, at $f = 0$, score 0.195 to 0.510 on two-hop questions, while the same pieces taken from one joint prefill score 0.985 to 1.000, a deficit of 48.5 to 79.5 points with $p < 1.3 \times 10^{-29}$ in every cell.

\begin{table}[h]
\caption{Threshold $\tau$ on two-hop questions, the fraction of the text that must be visible during preparation before the memory delivers 95\% of its benefit.}
\label{tab:tau}
\begin{center}
\begin{tabular}{lccccc}
\toprule
Text length (tokens) & 1{,}500 & 3{,}000 & 6{,}000 & 12{,}000 & 24{,}000 \\
\midrule
$\tau$, original run & 0.898 & 0.891 & 0.867 & 0.915 & 0.824 \\
$\tau$, fresh-item replication & 0.863 & 0.905 & 0.849 & 0.880 & 0.866 \\
\bottomrule
\end{tabular}
\end{center}
\end{table}

Single-hop questions behave differently. Accuracy at $f = 0$ is already 0.635 to 0.795 in the eight-piece configurations, and 0.48 to 0.97 over all, because the gold fact lives in a single paragraph and that paragraph's piece is present and intact in the composed memory. Visibility still adds 3 to 51 points depending on the configuration, the least where $f = 0$ starts near ceiling, and where the climb is large enough to fit a threshold, single-hop $\tau$ is 0.34 to 0.58, against 0.82 to 0.92 for two-hop. In 8 of 18 single-hop configurations the climb falls under the registered bar, and a robustness check with a different monotone fit reproduces the threshold in only half the configurations, so single-hop thresholds are directional estimates. A further directional check follows. If a two-hop item simply required two independent retrievals, two-hop accuracy would track the square of the single-hop curve, and it instead sits well below that square, with a mean residual of $-0.273$, so chaining two facts costs more than two independent retrievals would.

\paragraph{Reading.} Over the tested range (1{,}500--24{,}000 tokens) the requirement is a constant fraction. Reaching 95\% of fully prepared two-hop accuracy on a 24{,}000-token text means co-preparing roughly 20{,}000 tokens, which is most of the prefill the memory replaces, so preparation cannot be made cheap by making it partial, and the savings of reuse come only from paying the full preparation once and querying many times. A piece captured alone stores keys and values shaped only by itself, the relationships a two-hop question must traverse were never encoded, and no recombination at serving time restores them. The connection between the facts, not the facts themselves, is what co-preparation supplies. The results also show that crucial information is spread through the text rather than concentrated in a small core, since if some fifth of the material carried most of the cross-paragraph links, the threshold would sit far lower than $\sim$86\%. The tested lengths stop at 24{,}000 tokens, and the slope constrains nothing beyond them, where long-context degradation would also erode the full-prefill reference itself \citep{liu2024lost, hong2025context}. Section~\ref{sec:limits} returns to this boundary.

\section{Query time: repeated revisions to one fact}
\label{app:revisions}

The common case for a long-lived memory is not one correction but a history of them. An assignee changes three times in a quarter, a status flips repeatedly, and only the last version is current. This experiment measures that case, and it is the single place the method under test holds a structural advantage. When a fact has been revised several times, all versions are equally relevant to the question, so ordering updates by relevance cannot encode which is newest, while a merged memory keeps only the newest version by construction. The experiment is about caches, with text policies as their rivals, and no cartridge appears.

The experiment crosses pile sizes 32 and 256 with revision depths 1, 2, 3, and 5 and with 1, 4, or 16 simultaneously stale facts. Depth counts supersessions of one fact. At depth 3 the fact has changed three times, three versions sit in the pile, and only the last is current. Items follow the construction of Section~\ref{sec:corrections} and pass the same screen, so every accuracy below is a rescue rate on the same scale, and answers are forced choice. Text policies run at their tuned configurations, and the primary contrasts rest on 3{,}150 items, paired on item and pile size with no item dropped unpaired. Two questions fixed in advance govern reporting. The registered ordering rule asks whether the relevance policy, which cannot encode recency, decays over depth at least 5 points per unit depth faster than the newest-first policy, with a significant paired difference at depth 5. The registered merged-memory rule asks whether merged injection beats the best text policy by at least 10 points at depth 3 or more, judged against both the policy list frozen when the rule was written and that list extended with paste merged, because a win over the frozen list alone would credit the substrate for the keep-only-newest policy, which text can implement too. One planned condition, in which only the upstream fact of the chain is stale, was dropped before any data existed because it cannot be constructed. When the answer changes only indirectly there is no final update to place.

Both questions come back against the method. The ordering penalty is real and correctly signed but small. At depth 5 the newest-first policy rescues 0.197 against 0.155 for the relevance policy, a paired difference of $+4.24$ points ($p = 3.3 \times 10^{-6}$, past its MDE of 2.53), but the extra decay is about 1.6 points per revision, the difference of the two policies' least-squares decay slopes fitted over depths 1, 2, 3, and 5, a third of the size the rule was written to detect.
Merged injection never beats the same merge run as text: it does not beat the best text policy against the frozen list, and it loses against the list extended with paste merged, which runs the identical merge rule as text. The place precomputation had a principled reason to win is a place it does not win. The advantage belongs to the policy rather than the substrate, consistent with Section~\ref{sec:corrections}'s finding that captured updates lose the contest among many revisions faster than re-read text. Every rescue rate here is small, roughly 0.15 to 0.20 at depth 5, so these are rankings among policies that fail most of the time. For practical implementation, the merge itself is the valuable move. Keep only the newest version of each fact and serve it as text, and do not pay for injection to get it.

\section{How measurement moves these numbers}
\label{app:measure}

The three studies in this appendix exist because every number in this paper is a joint product of the model and the evaluator, and we wanted the evaluator's share measured rather than assumed.

\subsection{Evaluator choices}

We varied answer format between multiple choice and free-form, applied the wrong chat template that our own pinned model mirror ships, audited item identifier schemes, and re-measured a reported tie rate at scale. Answer format moves accuracy by 64 points on this instrument. The wrong template costs 4.76 points under multiple choice and 5.81 under free-form, pooled over 20 cells at $n = 3{,}991$, and 8.45 to 10.25 points on two-hop questions, and the bad template ships with a family of mirrors rather than one. Positional item identifiers produced 1{,}464 silent collisions across datasets before scoped identifiers replaced them, and a tie rate reported as 3.5\% at $n = 200$ narrowed to 0.195\% at $n = 55{,}874$. A published accuracy is a statement about a model and an evaluator together, and an automatic scorer validated on one answer format is not validated on another. These measurements are why the robustness grid in Section~\ref{sec:corrections} was run at all, and the template result is why the template is vendored and hash-checked at load.

\subsection{The honest ceiling}

Most gaps in this literature are quoted against a full-context read. On LongHealth \citep{adams2024longhealth}, 400 five-option questions over 20 fictional patients' records of 5{,}090 to 6{,}754 words (9{,}500 to 12{,}400 tokens under this paper's tokenizer) each, we measured the best trained cartridge across a budget sweep, and built a second evaluation whose items are selected independently of any strategy's performance, to compare against the usual practice of keeping the items the reference answers correctly. The full-context reference scores 23.0 points above the directly measured trained ceiling on the full set (0.780 against 0.550, $n = 200$) and 29.5 on the independently selected set (0.946 against 0.651, $n = 149$), the same cartridge on two item sets. The 0.651 is the sweep's primary 1{,}024-token budget on the independently selected items. The sweep's best cell, the same items at the 8{,}192-token budget, reaches 0.832 against a full-context 0.946, still 11.4 points short at only 1.3$\times$ compression, and neither is a statement about the best achievable cartridge.
A third selection, keeping the 156 items this instrument's own reference answers correctly, the usual practice, yields a 35.26-point gap against the 29.53 under independent selection. Selecting items on the reference's outcome therefore inflates the reported gap by 5.7 points, almost entirely in the ceiling. On the synthetic texts the full-context reference answers 98.5 to 100\% of items, so the hazard bites exactly when the reference is below ceiling, which is the situation on real records. Gaps quoted against full context are systematically generous, and the 23 to 29.5 point correction is used qualitatively rather than subtracted.

\subsection{Distractor confusability has two effective levels}

Three intended regimes, near-identical, same-type, and mixed-type, were crossed with both question tiers at all five text lengths, 200 items per cell and 47{,}928 measurements, pooled across lengths by design so that no favorable length could be chosen after the fact. The intended three-rung ladder has two rungs. Near-identical and same-type distractors are not distinguishable, at +3.83 and +5.52 points with neither significant, and all the difficulty comes from mixed-type being 18 to 47 points easier. The confusability penalty is larger for two-hop questions by 14.13 points under multiple choice, clearing a threshold fixed in advance of 10, but under free-form answers the same contrast is $-5.01$ and misses the threshold by about as much as it clears it under multiple choice. Both formats are reported and neither is chosen, a format disagreement of exactly the kind the first study in this appendix quantifies. Results measured on near-identical distractors may be described as applying to same-type distractors, because that equivalence is now measured, and they may not be extended to mixed-type.

\section{Precursor measurements}
\label{app:precursor}

The program's design drew on unpublished internal measurements from two precursor repositories, reported here for grounding. Both predate every rule in Appendix~\ref{app:setup} and carry no evidential weight in this paper's claims. The first set ran on other models, and the second on this paper's model but a different benchmark and judge, and we keep them apart so that neither is mistaken for a result.

\subsection{Parametric-pathway precursors on other models}

On a 35-item single-hop probe with Gemma-2-9b-it \citep{riviere2024gemma2}, one low-rank adapter trained per fact (rank 4, $\alpha = 16$, on the query and value projections) answered 6 of 35, against a no-memory floor of 12 of 35, a full-context read of 35 of 35, and a soft-token store of 29 of 35, trained virtual tokens at the input-embedding layer in the manner of prompt tuning \citep{lester2021power}. No trained key-value cartridge, the prefix-style object of Section~\ref{sec:conflict}, was in that probe. The floor exceeds nominal chance because 12 of the items are binary and the base model guesses them at roughly half, and the adapter's trained push to emit its stored sentence overrides those free correct guesses, which is how a parametric memory method lands below no memory at all. A later cross-family ladder on Qwen2.5-14B and Mistral-7B ordered the storage locations the same way: the text itself at 100\%, its prefilled cache at 97\%, the soft-token store at 83\%, LoRA at 40\%, and a trained key-value cartridge at 0\% under default initialization, a figure that proved to be an initialization artifact, since the same store reasons once initialized at a standard deviation of 0.02. The honest reading is that the key-value pathway reasons and the training objective and initialization are what kill it, the initialization sensitivity that Sections~\ref{sec:compose} and~\ref{sec:warmstart} meet again. Reasoning percentages from these early probes may carry instrument contamination that the recall numbers do not. Together, one-hop and 35 items, these measurements are why weight-pathway patching was set aside in favor of cache-level updating, alongside the published multi-hop propagation failures of parameter editing cited in Section~\ref{sec:conflict}.

\subsection{Cache-level precursors on this model}

A second precursor ran on Llama-3.1-8B-Instruct over the LoCoMo conversational benchmark \citep{maharana2024locomo}, under a different judge and before the rules of Appendix~\ref{app:setup}, so it is reported for grounding only. It maintained a bounded working set of five mutually conditioned key-value chunks, merging chunk pairs as material arrived, under three re-conditioning policies. Modularizing a co-loaded cache into five conditioned chunks cost no accuracy, 0.37 against 0.38 for the co-loaded cache and 0.29 for retrieval, recovering about nine-tenths of the co-load gap, and in a separate comparison the cheapest maintenance policy, conditioning each chunk once and never re-touching it, was also the most accurate, 0.404 against 0.372 for lazy re-conditioning and 0.365 for the co-loaded cache as measured there, with lazy re-conditioning about 3$\times$ cheaper than full re-conditioning. Its headline claim of beating retrieval did not survive a full-recall retrieval baseline, so what it shows is representation modularity and cheap maintenance rather than a per-query cost advantage, a third accumulation architecture beside the two priced in Section~\ref{sec:warmstart} and a down payment on the consolidation direction of Section~\ref{sec:discussion}. The same repository holds the direct ancestor of Section~\ref{sec:corrections}, a 36-item probe in which text beat injection and independent injection rescued nothing at two hops, the pattern Section~\ref{sec:corrections} measures at scale. It also holds the one replicated place where a conditioned injection beat every text policy, deep multi-hop update chains, where the conditioned arm stayed flat across depth while text collapsed and won 7 of 8 chain cells over two seeds, a nuance that does not weaken the verdict of Section~\ref{sec:corrections} but does feed its note that the conflict residue looks like a training target.
\end{document}

%% file: math_commands.tex
\usepackage{amsmath,amsfonts,bm}

\def\eqref#1{equation~\ref{#1}}

\def\1{\bm{1}}

\DeclareMathAlphabet{\mathsfit}{\encodingdefault}{\sfdefault}{m}{sl}
\SetMathAlphabet{\mathsfit}{bold}{\encodingdefault}{\sfdefault}{bx}{n}

